\documentclass{article}
\usepackage[margin=1in]{geometry}
\pdfoutput=1

\usepackage[textwidth=8em,textsize=small]{todonotes}

\usepackage{amsmath}
\usepackage{amssymb}
\usepackage{mathtools}

\usepackage{soul}

\usepackage{booktabs}

\usepackage{times}
\usepackage{bbm}

\usepackage{bm}
\usepackage{natbib}

\newtheorem{theorem}{Theorem}
\newtheorem{lemma}{Lemma}
\newtheorem{remark}{Remark}
\newtheorem{proof}{Proof}

\usepackage[caption=false, font=footnotesize]{subfig}

\usepackage{float}
\usepackage{placeins}

\usepackage{mathtools}

\usepackage{color}

\newcommand{\yh}[1]{\boldsymbol y^{[h]}_{#1}}

\newcommand{\tdhi}{\boldsymbol{\tilde d}^{[h]}_{i,1}}
\newcommand{\tdhii}{\boldsymbol{\tilde d}^{[h]}_{i,i'}}
\newcommand{\tdh}[1]{\boldsymbol{\tilde d}^{[h]}_{#1}}

\renewcommand{\dh}[1]{d^{[h]}_{#1}}

\newcommand{\dhii}{d^{[h]}_{i,i'}}

\newcommand{\be}{\begin{equation*}\begin{aligned}}

\newcommand{\ee}{\end{aligned}\end{equation*}}

\title{Bayesian Distance Clustering}

\author{
Leo L Duan\thanks{University of Florida, Gainesville, USA}
 \;\; and  David B Dunson\thanks{Duke University, Durham, USA}
}

\begin{document}

\date{}
\maketitle

\begin{abstract}
Model-based clustering is widely-used in a variety of application areas. However, fundamental concerns remain about robustness. In particular, results can be sensitive to the choice of kernel representing the within-cluster data density. Leveraging on properties of pairwise differences between data points, we propose a class of Bayesian distance clustering methods, which rely on modeling the likelihood of the pairwise distances in place of the original data.  Although some information in the data is discarded, we gain substantial robustness to modeling assumptions.  The proposed approach represents an appealing middle ground between distance- and model-based clustering, drawing advantages from each of these canonical approaches.  We illustrate dramatic gains in the ability to infer clusters that are not well represented by the usual choices of kernel.  A simulation study is included to assess performance relative to competitors, and we apply the approach to clustering of brain genome expression data.\\

\end{abstract}

\section{Introduction}

Clustering is a primary focus of many statistical analyses, providing a valuable tool for exploratory data analysis and simplification of complex data.  In the literature, there are two primary approaches -- distance- and model-based clustering.  Let ${\boldsymbol y}_i \in \mathcal{Y}$, for $i=1,\ldots,n$, denote the data and let $d(\boldsymbol y, \boldsymbol y')$ denote a distance between data points $\boldsymbol y$ and $\boldsymbol y'$.  Then, distance-based clustering algorithms are typically applied to the $n \times n$ matrix of pairwise distances
$D_{(n)\times (n)} = \{ d_{i,i'} \}$, with $d_{i,i'} = d(\boldsymbol y_i, \boldsymbol y_{i'})$ for all $i,i'$ pairs. For recent reviews, see \cite{jain2010data, xu2015comprehensive}. In contrast, model-based clustering takes a likelihood-based approach in building a model for the original data $\boldsymbol y_{(n)}$, with $(n) = \{ 1,\ldots,n \}$, that has the form:
\begin{eqnarray}
\boldsymbol y_i \stackrel{iid}{\sim} f,\quad f(\boldsymbol y) = \sum_{h=1}^k \pi_h \mathcal{K}(\boldsymbol y; \boldsymbol \theta_h), \label{eq:fmix}
\end{eqnarray}
where $\boldsymbol \pi = (\pi_1,\ldots,\pi_k)'$ is a vector of probability weights in a finite mixture model, $h$ is a cluster index, and $\mathcal{K}(\boldsymbol y; \boldsymbol\theta_h)$ is the density of the data within cluster $h$.  Typically, $\mathcal{K}(\boldsymbol y; \boldsymbol\theta)$ is a density in a parametric family, such as the Gaussian, with $\boldsymbol\theta$ denoting the parameters.  The finite mixture model (\ref{eq:fmix}) can be obtained by marginalizing out the cluster index $c_i \in \{1,\ldots,k\}$ in the following model:
\begin{eqnarray}
\boldsymbol y_i \sim \mathcal{K}( \boldsymbol\theta_{c_i} ),\quad \mbox{pr}(c_i=h) = \pi_h.  \label{eq:fmix2}
\end{eqnarray}
Using this data-augmented form, one can obtain maximum likelihood estimates of the model parameters $\boldsymbol\pi$ and $\boldsymbol\theta = \{\boldsymbol \theta_h\}$ via an expectation-maximization algorithm \citep{Fraley:2002bg}.  Alternatively, Bayesian methods are widely used to include prior information on the parameters, and characterize uncertainty in the parameters. {For recent reviews, see \cite{bouveyron2014model} and \cite{mcnicholas2016model}.}

Distance-based algorithms tend to have the advantage of being relatively simple conceptually and computationally,  while a key concern is the lack of characterization of uncertainty in clustering estimates and associated inferences. While model-based methods can address these concerns {by exploiting a likelihood-based framework}, a key disadvantage is large sensitivity to the choice of kernel $\mathcal{K}(\cdot; \boldsymbol \theta)$. Often, kernels are chosen for simplicity and computational convenience and they place rigid assumptions on the shape of the clusters, which are not justified by the applied setting being considered.

We are not the first to recognize this problem, and there is a literature attempting to address issues with kernel robustness in model-based clustering.  One direction is to choose a flexible class of kernels, which can characterize a wide variety of densities.  For example, one can replace the Gaussian kernel with one that accommodates asymmetry, skewness and/or heavier tails (\cite{karlis2009model,juarez2010model, o2016clustering, gallaugher2018finite}; among others).  A related direction is to nonparametrically estimate the kernels specific to each cluster, while placing minimal constraints for identifiability, such as unimodality and sufficiently light tails.  This direction is related to the mode-based clustering algorithms of \cite{li2007nonparametric}; see also \cite{rodriguez2014univariate} for a Bayesian approach using unimodal kernels.  Unfortunately,
{as discussed by \cite{hennig2015handbook}, a kernel that is too flexible leads to ambiguity in defining a cluster and identifiability issues: for example, one cluster can be the union of several clusters that are close. Practically, 
such flexible kernels demand a large number of parameters,  leading to daunting computation cost}. 

A promising new strategy is replace the likelihood with a robust alternative.   \cite{coretto2016robust} propose a pseudo-likelihood based approach for robust multivariate clustering, which {captures outliers with an extra improper uniform component}.  \cite{miller2018robust} propose a coarsened  Bayes approach for robustifying Bayesian inference and apply it to clustering problems.  Instead of assuming that the observed data are exactly generated from (\ref{eq:fmix}) in defining a Bayesian approach, they condition on the event that the empirical probability mass function of the observed data is within some small neighborhood of that for the assumed model. Both of these methods aim to allow small deviations from a simple kernel. It is difficult to extend these approaches to data with  high complexity, such as clustering multiple time series, images, etc.

We propose a new approach based on a Bayesian model for the pairwise distances, avoiding a complete specification of the likelihood function for the data $\boldsymbol y_{(n)}$.  There is a rich literature proposing Bayesian approaches that replace an exact likelihood function with some alternative.  \cite{chernozhukov2003mcmc} consider a broad class of such quasi-posterior distributions.  \cite{jeffreys1961theory} proposed a substitution likelihood for quantiles for use in Bayesian inference; refer also to \cite{dunson2005approximate}.  \cite{hoff2007extending} proposed a Bayesian approach to inference in copula models, which avoids specifying models for the marginal distributions via an extended rank likelihood. \cite{johnson2005bayes} proposed Bayesian tests based on modeling frequentist test statistics instead of the data directly. These are just some of many such examples.

Our proposed Bayesian Distance Clustering approach gains some of the advantages of model-based clustering, such as uncertainty quantification and flexibility, while significantly simplifying the model specification task.

\section{Partial likelihood for distances}

\subsection{Motivation for partial likelihood\label{sec:motivation}}

Suppose that data $\boldsymbol y_{(n)}$ are generated from model (\ref{eq:fmix}) or equivalently (\ref{eq:fmix2}).  
We focus on the case in which $\boldsymbol y_i = (y_{i,1},\ldots,y_{i,p})' \in \mathcal{Y} \subset \mathbb{R}^p$.
The conditional likelihood of the data $\boldsymbol y_{(n)}$ given clustering indices $\boldsymbol  c_{(n)}$ can be expressed as
\begin{eqnarray}
        L(\boldsymbol y_{(n)}; \boldsymbol  c_{(n)}) = \prod_{h=1}^k \prod_{i:c_i = h} \mathcal{K}_h( \boldsymbol y_i ) = \prod_{h=1}^k L_h( \boldsymbol y^{[h]} ), \label{eq:lik1}
\end{eqnarray}
where we let $ \mathcal{K}_h(\boldsymbol y)$ denote the density of data within cluster $h$, and $\boldsymbol y^{[h]} = \{ \boldsymbol y_i: c_i = h \} = \{ \yh{i}, i=1,\ldots,n_h \}$ is the data in cluster $h$. Since the information of $\boldsymbol c_{(n)}$  is stored by the index with $[h]$, we will omit $\boldsymbol c_{(n)}$ in the notation when $[h]$ appears. Referring to $\boldsymbol y^{[h]}_{1}$ as the {\em seed} for cluster $h$, we can express the likelihood $L_h(\boldsymbol y^{[h]})$ as
\begin{equation}
\label{eq:lik2}
    \begin{aligned}
L_h(\boldsymbol y^{[h]}  ) & =  \mathcal{K}_h ( \yh{1}  ) \;\prod_{i = 2}^{n_h}  G_h( \tdh{i,1}  \mid    \yh{1} )\\
& =\mathcal{K}_h \big( \yh{1}  \mid  \tdh{2,1}, \ldots ,\tdh{n_h,1}   \big)\; G_h \big( \tdh{2,1}, \ldots ,\tdh{n_h,1}\big), \\
\end{aligned}
\end{equation}
where $\tdhi$ denotes the difference between $\yh{i}$ and $\yh1$; {for now, let us define difference by subtraction $\tdhi  = \yh{i} - \yh1$}. Expression (\ref{eq:lik2}) is a product of the densities of the seed and $(n_h-1)$ differences. As the cluster size $n_h$ increases, the relative contribution of the seed density $\mathcal{K}_h( \yh{1}\mid. )$ will decrease and the likelihood becomes dominated by $G_h$. The term {$\mathcal{K}_h( \yh{1}\mid. )$} can intuitively be discarded with little impact on the inferences of $\boldsymbol  c_{(n)}$.

 Our interest is to utilize all the pairwise differences  ${\tilde D}^{[h]} = \{\tdh{i,i'}\}_{(i,i')}$, besides those formed with the seed. Fortunately, there is a relationship to determine the other  $\tdh{i,i'}=\tdh{i,1}-\tdh{i',1}$ for all $i'>1$. This means $\{\tdh{2,1},\ldots,\tdh{n_h,1}\}$, if affinely independent, is a minimal representation \citep{wainwright2003graphical} for ${\tilde D}^{[h]}$, which is the over-complete representation. As a result, one can induce a density for over-complete $\tilde D^{(h)}$ through a density for minimal $\{\tilde {\boldsymbol d}_{2,1}^{[h]},\ldots,\tilde  {\boldsymbol d}_{n_h,1}^{[h]}\}$ without restriction. We consider
\begin{equation}
\label{eq:prod_density}
        \begin{aligned}
   G_h \big( \tdh{2,1}, \ldots ,\tdh{n_h,1}\big) = \prod_{i=1}^{n_h}  \prod_{i'\neq i} g_h^{1/n_h}( \tdh{i,i'}),
        \end{aligned}
\end{equation}
where $g_h: \mathbb{R}^p\to \mathbb{R}_+$ and each $\tilde {\boldsymbol d}_{i,i'}^{[h]}$ is assigned a marginal density. To calibrate the effect of the over-completeness on the right hand side, we use a power  $1/n_h$, with this value justified in the next section.

From the assumption that data within a cluster are iid, we can immediately obtain two key properties of $\tdhii =\yh{i} - \yh{i'}$: (1) Expectation zero, and (2) Marginal  symmetry with skewness zero. 
 Hence, the distribution of the differences is  substantially simpler than the original data distribution $\mathcal{K}_h$. 
 This suggests using $G_h( \tilde D^{[h]})$ for clustering will substantially reduce the model complexity and improve robustness.


{
        We connect the density of the differences to a likelihood of `distances' ---  here used as a loose notion including metrics, semi-metrics and divergences. Consider $d_{i,i'}\in[0,\infty)$ as a transform of $\tilde {\boldsymbol d}_{i,i'}$, such as some norm  $ d_{i,i'}=\|\tilde {\boldsymbol d}_{i,i'}\|$ (e.g. Euclidean or $1$-norm); hence, a likelihood for $ d_{i,i'}$ is implicitly associated to a pushforward measure from the one on $\tilde {\boldsymbol d}_{i,i'}$ (assuming a measurable transform). For example, an exponential density on $d_{i,i'} = \| \tilde {\boldsymbol d}_{i,i'}\|_1$ can be taken as the result of assigning a multivariate Laplace on $\tilde {\boldsymbol d}_{i,i'}$. We can further generalize the notion of difference from substraction to other types, such as ratio, cross-entropy, or an application-driven specification \citep{izakian2015fuzzy}.

To summarize, this motivates the practice of first calculating a matrix of pairwise distances, and then assigning a partial likelihood for clustering.  For generality, we slightly abuse notation and replace difference array $\tilde D$ with distance matrix $D$ in \eqref{eq:prod_density}.
 We will refer to \eqref{eq:prod_density} as the {\em distance likelihood} from now on. Conditional on the clustering labels,
}

\begin{equation}
\label{eq:dp_likelihood}
        \begin{aligned}
L\{\boldsymbol y_{(n)} ; \boldsymbol  c_{(n)} \} =   \prod_{h=1}^k G_h(D^{[h]}),
        \end{aligned}
\end{equation}
with $c_i \sim \sum_{h=1}^k \pi_h \delta_h$ independently, as is (\ref{eq:fmix2}).

\subsection{Choosing a distance density for clustering}

\begin{figure}[H]
\begin{center}
\subfloat{}{
\includegraphics[width=1.5in]{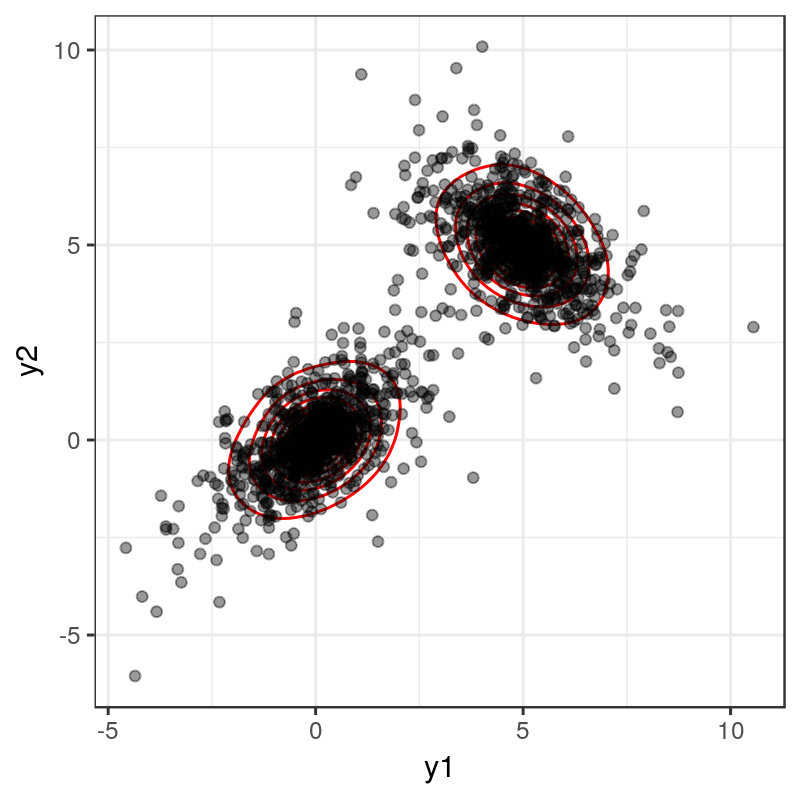}}
\hfill
\subfloat{}{\includegraphics[width=4in]{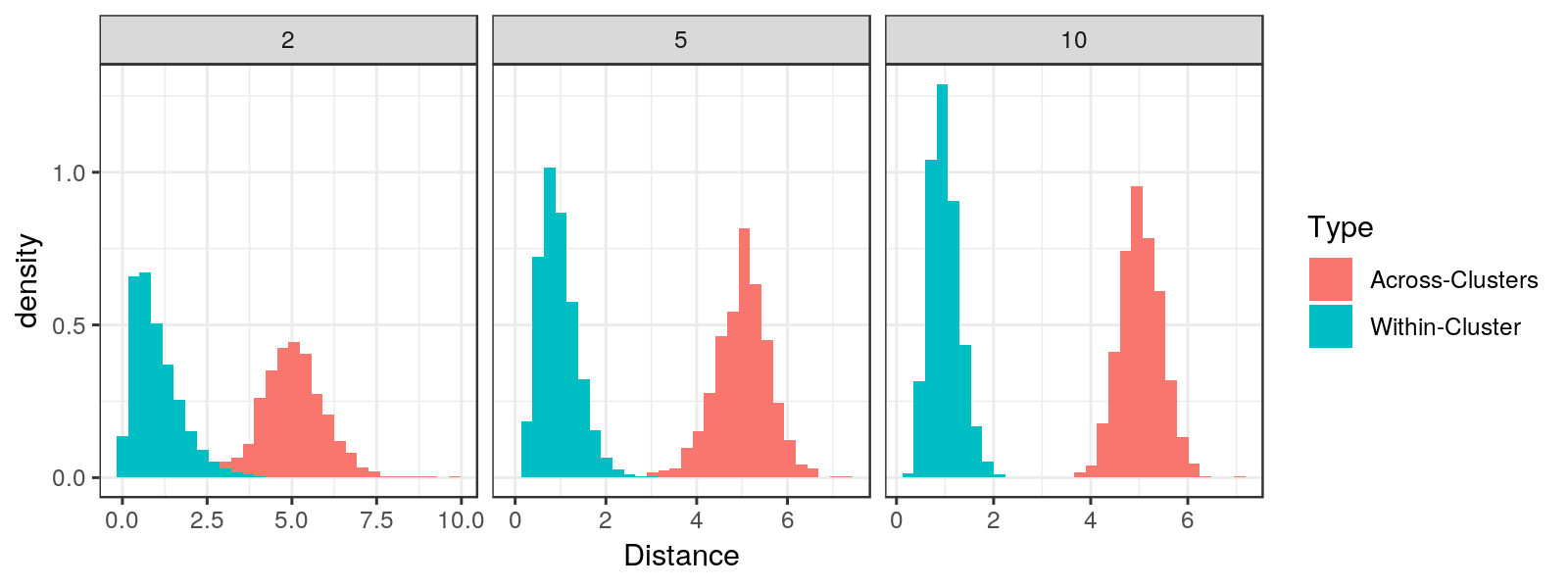}}
\caption{Histograms of  Euclidean distances scaled by $1/\sigma_h$ (with $\sigma_h\approx\sqrt{p}$). They show two characteristics of the distances formed within a cluster (cyan): 1. they tend to be much smaller than the ones across clusters (red); 2. as dimension $p$ increases (from $2$, $5$ to $10$), the mode of $\dh{i,i'}/\sigma_h$   becomes near $1$. Each cluster's data are generated from a multivariate Laplace distribution $y_{i} \sim \text{Lap}(\mu_h, \Sigma_{h})$ with $h=1,2$.
        \label{fig:concentration_vector_norm}}
\end{center}
\end{figure}

To implement our Bayesian distance clustering approach, we need a definition of clusters, guiding us to choose a parametric form for $g_h(.)$  in \eqref{eq:prod_density}.  A popular intuition for a cluster is a group of data points, such that most of the distances among them are relatively small.{ That is, the probability of finding large distances within a cluster should be low,
        \begin{equation}\label{eq:rapid_decline}
        \textup{pr} (  \dh{i,i'}  \ge  t \sigma_h) \le \epsilon_h(t) \;\;\text{ for sufficiently large }  t>0,
\end{equation}
with $ \sigma_h>0$ a scale parameter  and $\epsilon_h$ a function that rapidly declines
towards $0$ as $t$ increases. For such a decline, it is  common to consider exponential rate \citep{wainwright2019high}, $\epsilon_h(t)\approx\mathcal O\{\exp(-t)\}$.
 
In addition, we also want to model
the distribution of the small within-cluster distances accurately. 
As noted above, the distribution of the pairwise differences is automatically symmetric about zero. However, these differences $\tdh{i,i'}=\yh{i}-\yh{i'}$ are $p$-dimensional vectors, and we require a model instead for the density of $\dh{i,i'}=\|\tdh{i,i'}\|$.

To provide an intuition,  consider the asymptotic case with $p\to \infty$, under the assumption that  
$\tilde d^{[h]}_{i,i',j}$'s are iid for all $j$ (see \cite{biau2015high} for non-iid cases). By the law of large numbers, for $L_1$ distance,  $\dh{i,i'}/p = \sum_{j=1}^p | \tilde d^{[h]}_{i,i',j} |/p $ converges to $
\mathbb{E} | \tilde d^{[h]}_{i,i',j} |>0$; for Euclidean distance, 
$(\dh{i,i'})^2/p =\sum_{j=1}^p
| \tilde d^{[h]}_{i,i',j} |^2/p$ converges to $
\text{Var} (\tilde d^{[h]}_{i,i',j})>0$.  Therefore, as $p$ increases, with an  appropriately estimated $\sigma_h$ (hence having a proper order in $p$), most of $\dh{i,i'}/\sigma_h$  within a cluster will become closer and closer to a {\em  positive}  constant (which is also finite as long  as the above moments  exist). Figure~\ref{fig:concentration_vector_norm} provides an illustration using simulated Euclidean distances --- even at a small $p=5$, the scaled distance $\dh{i,i'}/\sigma_h$ starts to show a mode near one. Therefore, we need another parameter to accommodate the mode, in addition to the scale.
}

Motivated by the above discussion, we choose $g_h(.)$ in \eqref{eq:prod_density}  as Gamma $(\alpha_h,\sigma_h)$ with $\alpha_h \ge\ 1$.  This density has mode $(\alpha_h-1)\sigma_h \ge 0$. 
\begin{equation}
\label{eq:gamma_den}
        \begin{aligned}
                g_h(\dh{i,i'}) = \frac{1}{\Gamma(\alpha_{h})\sigma^{\alpha_h}_{h}}    x^{\alpha_{h}-1}\exp\left( -   \dh{i,i'}/{\sigma_{h} }\right).
        \end{aligned}
\end{equation}
We defer the prior choice for $\alpha_h$ and $\sigma_h$ to a later section. The following lemma provides a bound on the tail of the distance density \eqref{eq:gamma_den}.
 
 \begin{lemma}
(Bound on the right tail) 
If $d$ has the density \eqref{eq:gamma_den}, for any 
$\alpha_h \ge 1$ and $t>0$,
\be
 \text{pr}(d\ge t \sigma_{h})\le M
t^{\alpha_h} \exp{( - t)},
\ee
where   $M=\left({\alpha_h}\right)^{ -\alpha_h}
\exp(\alpha_h) $.
\end{lemma}

\begin{remark}
       The polynomial term $t^{\alpha_h}$ allows deviation from the exponential distribution at small $t$ (such as the non-zero mode); its effect vanishes as $t$ increases, giving  $\epsilon_h(t)\approx\mathcal O\{\exp(-t)\}$ as in \eqref{eq:rapid_decline}.
\end{remark}

\begin{remark}
One could imagine having more than one mode for the distances within a cluster. However, that often indicates that the cluster can be further broken into  smaller clusters.  As an example, consider the distance histogram in        Figure~\ref{fig:concentration_vector_norm}  with $p=2$: a bi-modal density $g_h(.)$ would under-fit the data, leading to a single large cluster. Therefore, for better separability of overlapping clusters, we use the single-mode density \eqref{eq:gamma_den} for $g_h(.)$ in expression  \eqref{eq:prod_density}.
\end{remark}

The assumptions on the density of the pairwise distances are connected to some implicit assumptions on the data distribution $\mathcal{K}(\boldsymbol y_i)$. As such a link varies with the specific form of the distance, we again focus on the vector norm of subtraction $\dhii = \|\yh{i} - \yh{i'}\|_q$, with $\|x\|_q =  (\sum_{j=1}^p x^q_j)^{1/q}$ and $q\ge 1$. We show in Theorem 1 that an exponential tail for the distribution of distances is a direct result of assuming sub-exponential tails in $\mathcal{K}(\boldsymbol y_i)$.

\begin{theorem}
\label{lm:concentration}
(Tail of vector norm distance) 
If there exist bound constants $m^{[h]}_1,m^{[h]}_2>0$, such that for all $j=1,\ldots,p$
\begin{equation}
\label{eq:concentration}
        \text{pr}(|y^{[h]}_{i,j} -\mathbb{E}y^{[h]}_{i,j} | \ge t)\le m^{[h]}_1 \exp(-m^{[h]}_2 t),
\end{equation} 
then, there exist another two constants $\nu_h, b_h>0$, such that
for any $q\ge 1$ 

\begin{equation}
\label{eq:sub_exp}
        \begin{aligned}
\text{pr}( \dh{ii'}>  tb_h p^{\eta}) \le 2 p \exp\{-t p^{(\eta-1/q)}/2
\}  \quad \text{  for  } t >  p^{1/q-\eta } 2\nu_h^2 .
        \end{aligned}
\end{equation}

\end{theorem}

\begin{remark}
The concentration property \eqref{eq:concentration} is less restrictive than common assumptions on the kernel in a mixture model, such as Gaussianity, log-concavity or unimodality.
\end{remark}

\section{Prior specification}

In Bayesian clustering it is useful to choose the prior parameters in a reasonable range \citep{malsiner2017identifying}. 
Recall in our gamma density, $\alpha_h \ge 1$ determines the mode for $\dh{i,i'}$ at $(\alpha_h-1)\sigma_h$. To favor small values for the mode while accommodating a moderate degree of uncertainty, we use a shifted Gamma prior $\alpha_h \sim \text{Gamma}(0.5, 1.0)+1$.

To select a prior for $\sigma_h$, we associate it with a pre-specified maximum cluster number $k$. We can view $k$ as a packing number --- that is, how many balls (clusters) we can fit in a container of the data.

To formalize, imagine a $p$-dimensional ellipsoid in $\mathbb{R}^p$ enclosing all the observed data. The smallest volume
of such an ellipsoid is\be
vol(\text{Data})= M{\min_{\mu\in \mathbb{R}^p,Q\succ 0}(\text{det } Q)^{-1/2}},  \; \text{ s.t. }    (\boldsymbol y_i- \boldsymbol \mu)^{\rm T} Q (\boldsymbol y_i- \boldsymbol\mu) \le 1 \text{ for } i =1,\ldots,n,
\ee
which can be obtained via a fast convex optimization algorithm \citep{sun2004computation}, with $M={ { \tilde \pi}^{p/2}}/{\Gamma(p/2+1)}$ and $\tilde \pi \approx 3.14$.

 If we view each cluster as a high-probability ball  of points originating from a common distribution, then the  diameter --- the distance between the two points that are  farthest apart --- is  $\sim 4\sigma_h$. This is calculated based on $\text{pr}(d\le 4\sigma_h)\approx 0.95$ using the gamma density with $\alpha_h=1.5$ (the prior mean of $\alpha_h$). We denote the ball by $\mathcal B_{2\sigma_h }$, with $vol(\mathcal B_{2\sigma_h })=M ({2\sigma_h })^p$. 

Setting $k$ to the packing number
\be
k \simeq \frac{vol(\text{Data})}{ vol(\mathcal B_{2\sigma_h })}
\ee
yields a sensible prior mean for $\sigma_h$. For conjugacy, we choose an inverse-gamma prior for $\sigma_h$ with $\mathbb{E}(\sigma_h)=\beta_h,$
\be
\sigma_h \sim \text{Inverse-Gamma}(2, \beta_{\sigma}), \qquad
 \beta_{\sigma} =  \frac{1}{2} \bigg\{ \frac{vol(\text{Data}) }{kM }  \bigg\}^{1/p}.
\ee

The above prior can be used as a default in broad applications, and does not
require tuning to each new application.

\section{Properties}
We describe several interesting properties for the distance likelihood.
\begin{theorem}
    {\em(Exchangeability)}
    When the product density \eqref{eq:prod_density} is used for all $G_h(D^{[h]})$, $h=1,\ldots,k$, the  distance likelihood \eqref{eq:dp_likelihood} is invariant to  permutations of the indices $i$:
$$   L \{  \boldsymbol y_{(n)} ; \boldsymbol c_{(n)} \}  = L\{ \boldsymbol y_{(n^*)} ; \boldsymbol c_{(n^*)} \}, $$
with $(n^*)=\{1_*,\ldots, n_*\}$ denoting a set of permuted indices.
\end{theorem}
\begin{remark}
  Under this exchangeability property, selecting different seeds does not change the distance likelihood.
\end{remark}

 We fill a missing gap between the model-based and distance likelihoods through considering an information-theoretic analysis of the two clustering approaches.  This also leads to a principled choice of the power $1/n_h$ in \eqref{eq:prod_density}.

To quantify the information in clustering, we first briefly review the concept of Bregman divergence \citep{bregman1967relaxation}. Letting $\phi: \mathcal{S}\to \mathbb{R}$ be a strictly convex and differentiable function, with $\mathcal{S}$ the domain of $\phi$, the Bregman divergence is defined as
\begin{equation*}
        \begin{aligned}
    B_{\phi}(\boldsymbol x,\boldsymbol y) = \phi(\boldsymbol x) - \phi(\boldsymbol y) - (\boldsymbol x- \boldsymbol y)^{\rm T} \triangledown \phi(\boldsymbol y),
        \end{aligned}
\end{equation*}
where $\triangledown \phi( \boldsymbol y)$ denotes the gradient of $\phi$ at $\boldsymbol y$. A large family of loss functions, such as squared norm and Kullback-Leibler divergence, are special cases of the Bregman divergence with suitable $\phi$. For model-based clustering, when the regular exponential family (`regular' as the parameter space is a non-empty open set) is used for the component kernel $\mathcal{K}_h$,  \cite{banerjee2005clustering} show that there always exists a re-parameterization of the kernel using Bregman divergence. Using our notation,
\begin{equation*}
        \begin{aligned}
    \mathcal{K}_h(\boldsymbol y_i; \boldsymbol \theta_h)=\exp\left\{  T(\boldsymbol y_i)' \boldsymbol\theta_h - \psi(\boldsymbol \theta_h) \right\} \kappa(\boldsymbol y_i) \Leftrightarrow \exp\left[ -B_{\phi} \left\{ T(\boldsymbol y_i),\boldsymbol \mu_h \right\}\right] b_{\phi}\{ T(\boldsymbol y_i)\},
        \end{aligned}
\end{equation*}
where $T(\boldsymbol y_i)$ is a transformation of $\boldsymbol y_i$, in the same form as the minimum sufficient statistic for $\boldsymbol \theta_h$ (except this `statistic' is based on only one data point $\boldsymbol y_i$);  $\boldsymbol \mu_h$ is the expectation of $T(\boldsymbol y_i)$ taken with respect to $\mathcal{K}_h(\boldsymbol y;\boldsymbol \theta_h)$; $\psi$, $ \kappa$ and $b_{\phi}$ are functions mapping to $(0,\infty)$.

With this re-parameterization, maximizing the model-based likelihood over $\boldsymbol c_{(n)}$ becomes equivalent to minimizing the within-cluster Bregman divergence
\begin{equation*}
        \begin{aligned}
     H_y = \sum_{h=1}^{k} H_{y}^{[h]}, \quad H_{y}^{[h]}=\sum_{i=1}^{n_h} B_{\phi} 
     \left \{ T(\yh{i}),\boldsymbol \mu_h \right\}.
        \end{aligned}
\end{equation*}
We will refer to $H_y$ as the model-based divergence.

For the distance likelihood, {considering those distances that can be viewed or re-parameterized as a pairwise Bregman divergence}, we assume each $g(\dhii)$ in the distance likelihood \eqref{eq:prod_density} can be re-written with a calibrating power $\beta_h>0$ as
\begin{equation*}
        \begin{aligned}
    g^{\beta_h} (\dhii)= z^{\beta_h}\exp \left[ -  \beta_h B_\phi \left\{  T(\yh{i}), T(\yh{i'}) \right \} \right],
        \end{aligned}
\end{equation*}
with $z>0$  the normalizing constant. A distance-based  divergence $H_d$ can be computed
as\begin{equation}
\label{eq:dist_bregman}
        \begin{aligned}
                H_d = \sum_{h=1}^{k} H_{d}^{[h]}, \quad H_{d}^{[h]}=\beta_h\sum_{i=1}^{n_h} \sum_{i'=1}^{n_h}  \frac{1}{2} B_\phi \left\{ T(\yh{i}), T(\yh{i'}) \right\}.
        \end{aligned}
\end{equation}

We now compare these two divergences $H_y$ and $H_d$ at their expectations.
        \begin{theorem}
(Expected Bregman Divergence) The distance-based Bregman divergence \eqref{eq:dist_bregman} in cluster $h$ has
\begin{equation*}
        \begin{aligned}
& \mathbb{E}_{\boldsymbol y^{[h]}} H_{d}^{[h]}=  \beta_h \mathbb{E}_{\boldsymbol y^{[h]}_i} \mathbb{E}_{\boldsymbol y^{[h]}_{i'}}  \sum_{i=1}^{n_h} \sum_{i'=1}^{n_h}  \frac{1}{2} B_\phi \{ T(\yh{i}), T(\yh{i'})\} \\
&=    ( n_h\beta_h) \; \mathbb{E}_{\boldsymbol y^{[h]}} \; \left[ \sum_{i =1}^{n_h}  \frac{B_\phi \{ T(\yh{i}), \boldsymbol \mu_h\} + B_\phi\{\boldsymbol \mu_h,T(\yh{i})\} }{2} \right],
        \end{aligned}
\end{equation*}
where the expectation over ${\boldsymbol y^{[h]}}$ is taken with respect to $ \mathcal{K}_h$.
\end{theorem}
\begin{remark}
The term inside the expectation on the right hand side is the symmetrized Bregman divergence between $T(\yh{i})$ and $\boldsymbol \mu_h$, which is close to $B_\phi \{ T(\yh{i}), \boldsymbol \mu_h\}$ in general \citep{banerjee2005clustering}. Therefore,  $ \mathbb{E}_{\boldsymbol y^{[h]}} H_{d}^{[h]} \approx ( n_h\beta_h )\mathbb{E}_{\boldsymbol y^{[h]}} H_{y}^{[h]}$; equality holds exactly if $B_\phi(\cdot,\cdot)$ is  symmetric.
\end{remark}

There is an order difference $\mathcal{O}(n_h)$ between distance-based and model-based divergences. Therefore, a sensible choice is simply setting $\beta_h=1/n_h$. This power is related to the weights used in composite pairwise likelihood \citep{lindsay1988composite, cox2004note}.

It is also interesting to consider the matrix form of the distance likelihood. We use $C$ as an $n\times k$ binary matrix encoding the cluster assignment,  
with $C_{i,h}=1$ if $c_i=h$, and all other $C_{i,h'}=0$. Then it can be verified that $C^{\rm T}C= \text{diag}(n_1,\ldots,n_h)$. Hence the distance likelihood, with the Gamma density, is
\begin{equation}
\label{eq:dist_like_mat}
       G(D) \propto \exp \big[\text{tr}\big\{C^{\rm{T}}
        (\log D) C  \Lambda {(C^{\rm{T}}C)^{-1}}\big\}\big]\exp\big[- \text{tr}  \big\{ C^{\rm T} D C   \big (\Sigma C^{\rm T}  C  \big)^{-1} \big\}\big],
\end{equation}
where $D$ is the $n\times n$ distance matrix, $\log$ is applied element-wise,  $\Sigma=\text{diag}(\sigma_1,\ldots,\sigma_k)$, and $\Lambda=\text{diag}(\alpha_1-1,\ldots,\alpha_h-1)$. If $C$ contains zero columns, the inverse is replaced by a generalized inverse. 

One may notice some resemblance of \eqref{eq:dist_like_mat} to the loss function in graph partitioning algorithms. Indeed, if we simplify the parameters to $\alpha_1=\cdots = \alpha_k=\alpha_0$ and $\sigma_1=\cdots = \sigma_k=\sigma_0$,
then
\begin{equation}
    \label{eq:dist_like_mat1}
   G(D)\propto \exp \big[\text{tr}  \big\{ C^{\rm T} A C   \big ( C^{\rm T}  C  \big)^{-1}\big\}\big],
\end{equation}
where $A=\kappa {\bf 1}_{n,n}-D/\sigma_0+ (\alpha_0-1)\log D$ can be considered as an adjacency matrix of a graph formed by a log-Gamma distance kernel, with ${\bf 1}_{n,n}$ as an $n\times n$ matrix with all elements equal to $1$; $\kappa$ a constant so that each $A_{i,j}>0$ (since $\kappa$ enters the likelihood as a constant $\text{tr} \{ C^{\rm T}\kappa {\bf 1}_{n,n}C(C^{\rm T}C)^{-1}\}=n\kappa$, it does not impact the likelihood of $C$). To compare, the popular normalized graph-cut loss \citep{bandeira2013cheeger} is
\begin{equation}
\label{eq:graph_cut}
        \text{NCut-Loss} = \sum_{h=1}^k\sum_{i: c_i=h} \sum_{j: c_j \neq h} \frac{A_{i,j} }{2n_h},
\end{equation}
which is the total edges deleted because of partitioning (weighted by $n_h^{-1}$ to prevent trivial cuts). There is an interesting link between \eqref{eq:dist_like_mat1} and \eqref{eq:graph_cut}.

\begin{theorem}
        Considering a graph with weighted adjacency matrix $A$,
the normalized graph-cut loss is related to the negative log-likelihood (omitting constant) \eqref{eq:dist_like_mat1}  via
\begin{equation*}
 2 \textup{NCut-Loss}         =
-\text{tr}  \big\{ C^{\rm T} A C   \big ( C^{\rm T}  C  \big)^{-1}\big\} +
 \sum_{i=1}^n \frac{ \sum_{j=1}^n A_{i,j}}{n_{c_i}}.
\end{equation*}
\end{theorem}
\begin{remark}
The difference on the right is often known as the degree-based regularization (with $\sum_{j=1}^n A_{i,j} $ the degree, $n_{c_i}$ the size of the cluster that data  $i$ is assigned to). When the cluster sizes are relatively balanced, we can ignore its effect.
Such a near-equivalence suggests that we can exploit popular graph clustering algorithms, such as spectral clustering, for good initiation of $C$ before posterior sampling.
In addition, our proposed method effectively provides an approach for uncertainty quantification in  normalized graph cuts.\end{remark}

\section{Posterior computation}
\label{sec:computation}

Since the likelihood is formed in a pairwise way, updating one cluster assignment $c_i$ has an impact on the others. Therefore, conventional Gibbs samplers updating one $c_i$ at a time are not ideal. Instead, we consider Hamiltonian Monte Carlo (HMC) to update the whole binary matrix $C$ in \eqref{eq:dist_like_mat}, via a lift-and-project strategy \citep{balas2002lift}: first lift each row $\boldsymbol C_i$ (on a simplex vertex) into the interior of the simplex, and denote the relaxation by $\boldsymbol W_i \in \Delta^{(k-1)}_{\setminus \partial}$, run leap-frog updates and then project it back to the nearest vertex as a proposal. We call this algorithm  lift-and-project HMC. This iterates in the following steps:
\begin{enumerate}
        \item Initialize each $\boldsymbol W_i$ and sample momentum variable $Q \in \mathbb{R}^{n\times k}$, with each $q_{i,j}\sim \text{No}(0, \sigma^2_q)$.
        \item  Run the leap-frog algorithm in the simplex interior $\Delta^{(k-1)} _{\setminus \partial} $ for $L$ steps using kinetic and potential functions:
                \be
                & K(Q) = \frac{1}{2 \sigma^2_q} \text{tr}( Q^{\rm T}Q) \\
                & U(W) = 
                - \text{tr}\big\{W^{\rm{T}} (\text{log} D)
      W  \Lambda {(W^{\rm{T}}W)^{-1}}\big\} 
        + \text{tr}  \big\{ W^{\rm T} DW   \big (\Sigma W^{\rm T}  W \big)^{-1} \big\},
                \ee
                with $W$ the matrix of $\{\boldsymbol W_i\}$.
                Denote the last state by $(W^*,Q^*)$.
\item Compute vertex projection $\boldsymbol C^*_i$ by setting the largest coordinate in $\boldsymbol W_i$ to $1$ and others to $0$ (corresponding to minimizing Hellinger distance between $\boldsymbol C^*_i$ and $\boldsymbol W_i$).
        \item Run Metropolis-Hastings, and accept proposal $C^{*}$ if
                \be
                u <\frac{U(C^*)K(Q^*)}{U(C)K(Q)},
                \ee
                where $u\sim \text{Uniform}(0,1)$.
    \item Sample
            $$ \sigma_{h}\sim\text{Inverse-Gamma} \bigg\{ \frac{(n_h-1)}{2}+2, \frac {\sum_{i,i'}\dh{i,i'}} {2  n_h}+ \beta_{\sigma}\bigg\} ,$$
      if $n_h>1$; otherwise update from $\sigma_h\sim\text{\text{Inverse-Gamma}} (2, \beta_{\sigma}).$
        \item Sample $(\pi_1,\ldots,\pi_h) \sim \text{Dir}(\alpha+n_1, \ldots,\alpha+n_h)$.
        \item Sample $\alpha_h$ using random-walk Metropolis.
\end{enumerate}
To prevent a negative impact of the projection on the acceptance rate, we make $\boldsymbol W_i$ very close to its vertex projection $\boldsymbol C^*_i$. This is done via a tempered softmax re-parameterization \citep{maddison2016concrete}
\be
w_{i,h} = \frac{\exp(v_{i,h}/t)}{\sum_{h'=1}^k \exp(v_{i,h'}/t)}, \quad h=1,\ldots,k.
\ee
At small $t$, if one $v_{i,h}$ is slightly larger than the rest in $\{v_{i,1}, \ldots, v_{i,k}\}$, then $w_{i,h}$ will be close to $1$. In this article, we choose $t=0.1$ and prevent the leap-frog stepsize from being too small, so that during Hamiltonian updates, each $\boldsymbol W_i$ is very close to a vertex with high probability.

We utilize the auto-differentiation toolbox in Tensorflow. To produce a point estimate of $\hat {\boldsymbol c}_{(n)}$, we minimize the variation of information as the loss function, using the algorithm provided in \cite{wade2018bayesian}. {
In the posterior computation of clustering models, a particular challenge has been the label-switching issue due to the likelihood equivalence when permuting the labels $h\in \{1,\ldots,k\}$. This often leads to  difficulties in diagnosing  convergence and assessing the assignment probability $\text{pr}(c_i=h)$. To completely circumvent this issue, we instead track the matrix product $CC^{\rm T}$ for convergence, which is invariant to the permutation of the columns of $C$ (corresponding to the labels). The posterior samples of  $CC^{\rm T}$  give an estimate of the pairwise co-assignment probabilities $\text{pr}(c_i=c_{i'}) = \sum_{h=1}^k \text{pr}(c_i=c_{i'}=h)$. To obtain estimates for $\text{pr}(c_i=h)$, we use  symmetric simplex matrix factorization \citep{duan2019latent} on $\{ \text{pr}(c_i=c_{i'})\}_{i,i'}$ to obtain an $n\times k$ matrix corresponding to $\{\text{pr}(c_i=h)\}_{i,h}$. The factorization can be done almost instantaneously.
}

\section{Numeric experiments}
\subsection{Clustering with skewness-robust distance}

As described in Section \ref{sec:motivation}, the vector norm based distance is automatically robust to skewness. To illustrate,
 we generate  $n=200$ data from a two-component mixture of skewed Gaussians:
\begin{equation*}
        \begin{aligned}
        & \text{pr}(c_i=1)=\text{pr}(c_i=2)=0.5,  \\
    & y_{i,j} \mid c_i=h \sim \text{SN}(\mu_{h},1,\alpha_{h})\text{ for } j=1\ldots p,
        \end{aligned}
\end{equation*}
where $\text{SN}(\mu,\sigma,\alpha)$ has density $\pi(y\mid \mu,\sigma,\alpha)= 2 f\{ (y-\mu)/\sigma \}F \{ \alpha(y-\mu)/\sigma \}$ with $f$ and $F$ the density and cumulative distribution functions for the standard Gaussian distribution.

\begin{figure}[H]
\captionsetup[subfloat]{width=2.5in}
\begin{center}
\subfloat[Histogram and the true density (red line) of a mixture of two symmetric Gaussians.]{}{
      \includegraphics[width=2.5in, height=1.5in]{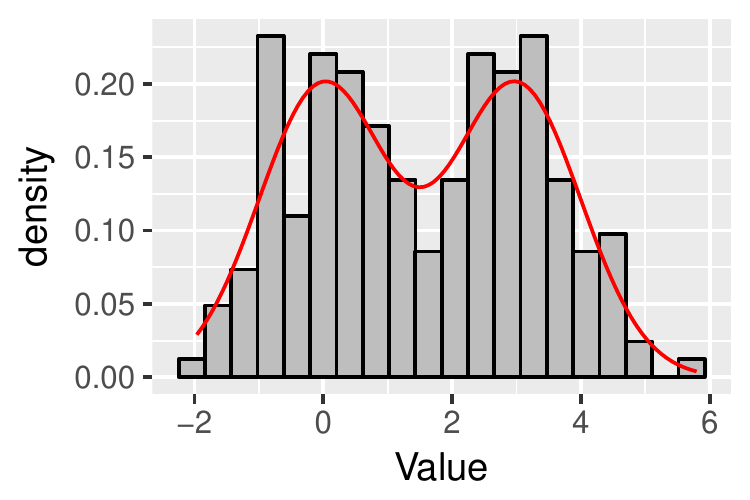}
}
\hfill
\subfloat[Histogram and the true density (red line) of a mixture of two right skewed Gaussians.]{}{
      \includegraphics[width=2.5in, height=1.5in]{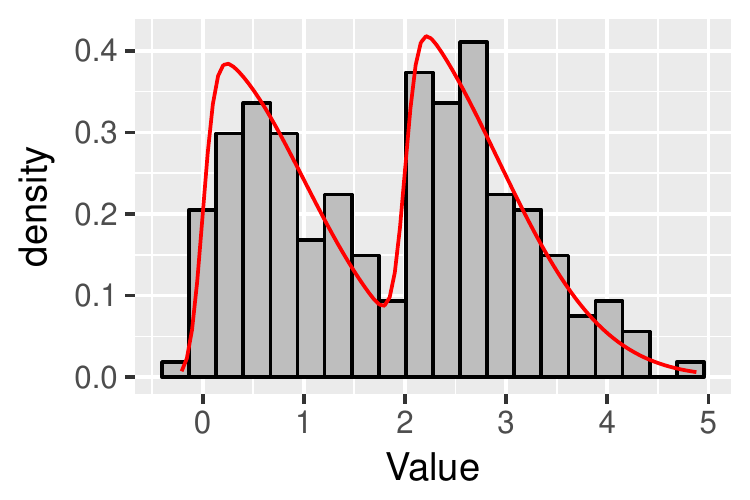}
}\\
\subfloat[Assignment  probability $\text{pr}(c_i=1)$, under Bayesian distance clustering and the mixture of Gaussians. Dashed line is the oracle probability based on symmetric Gaussians.]{}{
      \includegraphics[width=2.5in, height=1.5in]{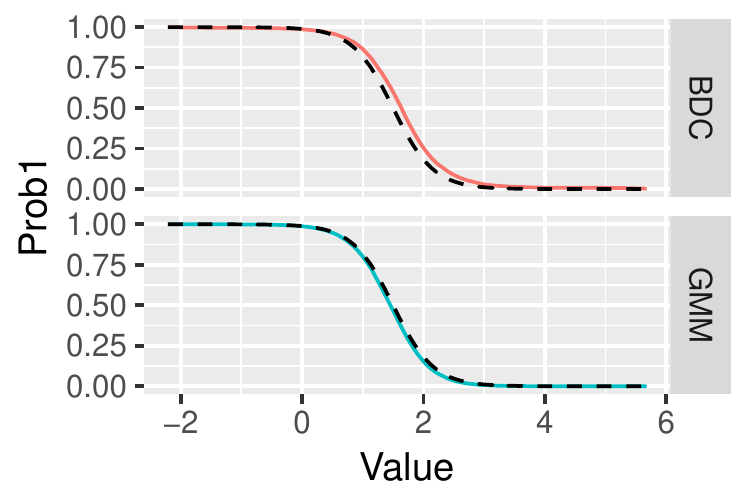}
}
\hfill
\subfloat[Assignment  probability $\text{pr}(c_i=1)$, under Bayesian distance clustering and the mixture of Gaussians. Dashed line is the oracle probability based on skewed Gaussians.]{}{
      \includegraphics[width=2.5in, height=1.5in]{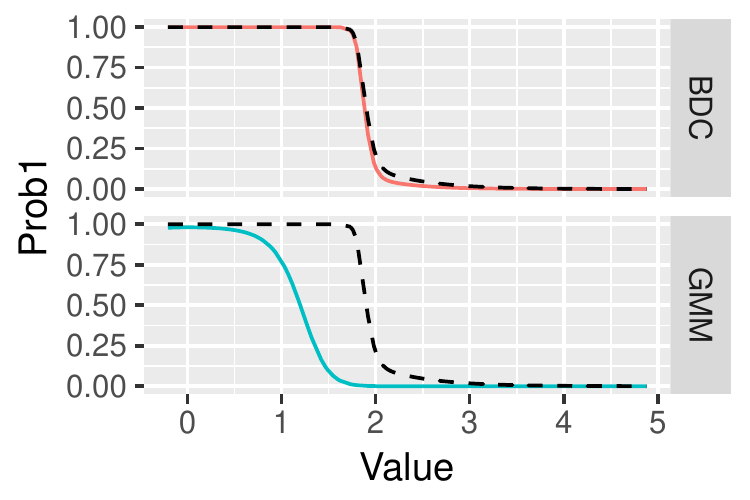}
}
\end{center}
  \caption{Clustering data from a two component mixture of skewed Gaussians in $\mathbb{R}$. Bayesian Distance clustering (BDC) gives posterior clustering probabilities close to the oracle probabilities
  regardless of whether the distribution is skewed or not (upper plots in panel c and d), while the mixture of Gaussians fails when the skewness is present (lower plot in panel d).\label{plt:gmm_simulations}}
\end{figure}

We start with $p=1$ and assess the performance of the Bayesian distance clustering model under both non-skewed ($\alpha_1=\alpha_2=0,\mu_1=0, \mu_2=3$) and skewed distributions ($\alpha_1=8, \alpha_2=10,\mu_1=0, \mu_2=2$). The results are compared against the mixture of Gaussians as implemented in the {\em Mclust} package. Figure~\ref{plt:gmm_simulations}(a,c) show that for non-skewed Gaussians, the proposed approach produces clustering probabilities close to their oracle probabilities, obtained using knowledge of the true kernels that generated the data. When the true kernels are skewed Gaussians, Figure~\ref{plt:gmm_simulations}(b,d) shows that the mixture of Gaussians gives inaccurate estimates of the clustering probability, whereas Bayesian distance clustering remains similar to the oracle.

To evaluate the accuracy of the point estimate $\hat c_i$, we compute the adjusted Rand index \citep{rand1971objective} with respect to the true labels. We test under different $p\in \{1,5,10,30\}$, and repeat each experiment $30$ times. The results are compared against model-based clustering using symmetric and  skewed Gaussians kernels, using independent variance structure. As shown in Table~\ref{tab:sim_ari_skewness},  the misspecified symmetric model deteriorates quickly as $p$ increases. 
In contrast, Bayesian distance clustering maintains high clustering accuracy.

\begin{table}
\caption{Accuracy of clustering  skewed Gaussians under different dimensions $p$. Adjusted Rand index (ARI) is computed for the point estimates using variation of information. The average and $95\%$ confidence interval are shown.\label{tab:sim_ari_skewness}}
\centering
    \begin{tabular}{l| c c c  }
                  \hline
  $p$           & Bayes Dist. Clustering & Mix. of Gaussians & Mix. of Skewed Gaussians\\
\hline
 1 &  0.80 (0.75, 0.84) &  0.65 (0.55, 0.71) & 0.81 (0.75, 0.85)\\
 5 &  0.76 (0.71, 0.81) &  0.55 (0.40, 0.61) & 0.76 (0.72, 0.80)\\
 10 &  0.72 (0.68, 0.76) &  0.33(0.25, 0.46) & 0.62 (0.53, 0.71)\\
 30 &  0.71 (0.67, 0.76) &  0.25 (0.20, 0.30) & 0.43 (0.37, 0.50)\\
\hline
\end{tabular}
\end{table}

\subsection{Clustering with subspace distance}

For high dimensional clustering, it is often useful to impose the additional assumption that each cluster lives near a different low-dimensional manifold. Clustering data based on these manifolds is known as subspace clustering. We  exploit the sparse subspace embedding algorithm proposed by \cite{vidal2013subspace} to learn pairwise subspace distances.
 Briefly speaking, since the data in the same cluster are alike, each data point can be approximated as a linear combination of several other data points in the same subspace; hence a sparse locally linear embedding can be used to estimate an $n\times n$ coefficient matrix $\hat W$ through
\be
\hat W =\arg\min_{W: w_{i,i}=0, \sum_j w_{i,j}=1}\sum_{i=1}^n\|\boldsymbol y_i - W \boldsymbol  y_i\|^2_2+ \|W\|_1,
\ee
where the sparsity of $\hat W$ ensures only the data in the same linear subspace can have non-zero embedding coefficients. Afterwards, we can define a subspace distance matrix as 
\[
d_{i,j} = 2-\bigg(\frac{| \hat  w_{i,j}|}{\max_{j'}|\hat  w_{i,j'}|}+\frac{|\hat  w_{j,i}|}{\max_{i'}|\hat w_{j,i'}|}\bigg),
\]
where we follow  \cite{vidal2013subspace} to normalize each row by its absolute maximum. We then use this distance matrix in our Bayesian distance clustering method.

To assess the performance, we use the MNIST data of hand-written digits of $0-9$, with each image having $p=28\times
28$ pixels. In each experiment, we take $n=5,000$ random samples to fit the clustering models, among which each digit has approximately $500$ samples, and we repeat experiments $10$ times. For comparison, we also run the near low-rank mixture model in {\em HDclassif} package \citep{berge2012hdclassif} and spectral clustering  based on the $p$-dimensional vector norm. Our method using subspace distances shows clearly higher accuracy as shown in Table~\ref{tab:mnist}.
\begin{table}
\caption{Accuracy of clustering MNIST hand-written digit data. Adjusted Rand index (ARI) is computed for the
point estimates using variation of information. The
average ARI and $95\%$ confidence intervals are shown.\label{tab:mnist}}
\centering
    \begin{tabular}{ c c c  }
                  \hline
  $$            Bayes Dist. Clustering & Spectral Clustering & HDClassif
\\
\hline
   0.45 (0.41, 0.52) &  0.31 (0.23, 0.35) & 0.35 (0.31, 0.43)\\
\hline
\end{tabular}
\end{table}

\subsection{Clustering constrained data}

In model-based clustering, if the data are discrete or in a constrained space, one would use a distribution customized to the type of data. For example, one may use the multinomial distribution for categorical data, or the directional distribution \citep{khatri1977mises} for data on a unit sphere. Comparatively,  distance clustering is simpler to use. {We can choose an appropriate distance for constrained data, and then use the same Bayesian distance clustering method as proposed.}

We consider clustering data on the unit sphere $\mathbb{S}^{p-1}=\{\boldsymbol  y:\boldsymbol y\in\mathbb{R}^p,\|\boldsymbol  y\|_2=1\}$ and generate $n=400$ data from a two component von-Mises Fisher (vMF) mixture:
\begin{equation*}
        \begin{aligned}
    \boldsymbol  y_{i} \sim 0.5  \; \text{vMF}(\boldsymbol \mu_{1},\kappa_1) + 0.5\; \text{vMF}(\boldsymbol \mu_{2}, \kappa_2) ,
        \end{aligned}
\end{equation*}
where $\boldsymbol  y\sim \text{vMF}(\boldsymbol  \mu,\kappa)$ has density proportional to $\exp(\kappa \boldsymbol \mu^{\rm T}\boldsymbol  y)$, with $\|\boldsymbol  \mu\|_2=1$. We present results for $p=2$, but similar conclusions hold for $p>2$. We fix $\kappa_1 = 0.25$, $\kappa_2 = 0.3$ and $\boldsymbol  \mu_1=(1,0)$, and vary $\boldsymbol  \mu_2$ for different separation between the two clusters. We measure the separation via the length of the arc between $\boldsymbol \mu_1$ and $\boldsymbol \mu_2$.   In this example, we use the absolute $\arccos$ distance between two points $d_{i,j}=|\cos^{-1} (\boldsymbol y_i^{\rm T} \boldsymbol y_j)|$

\begin{figure}[H]
\captionsetup[subfloat]{width=1.6in}
\begin{center}
\subfloat[Data on unit circle colored by true cluster labels.]{}{
      \includegraphics[width=1.8in, height=1.5in]{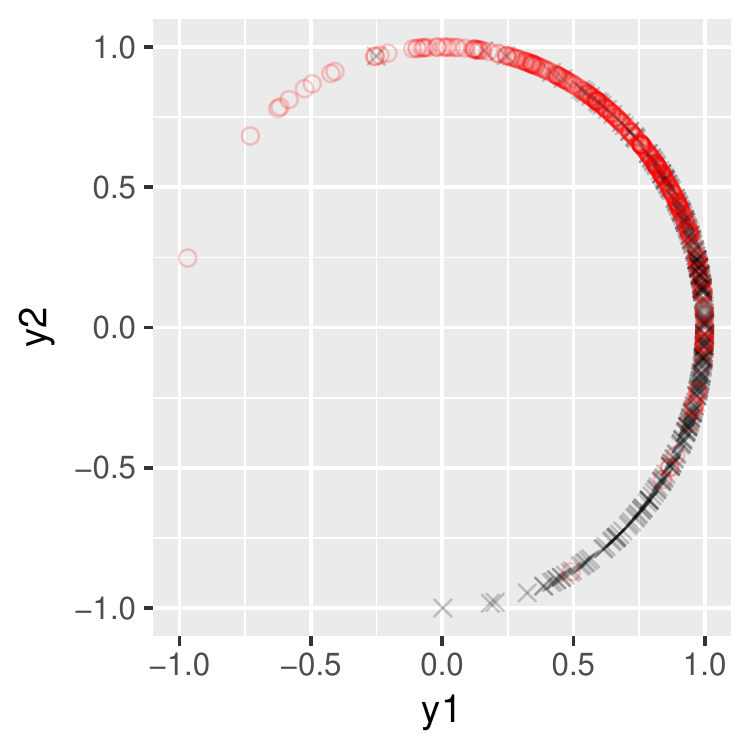}
}
\hfill
\subfloat[Point clustering estimates from a mixture of Gaussian model.]{}{
      \includegraphics[width=1.5in, height=1.5in]{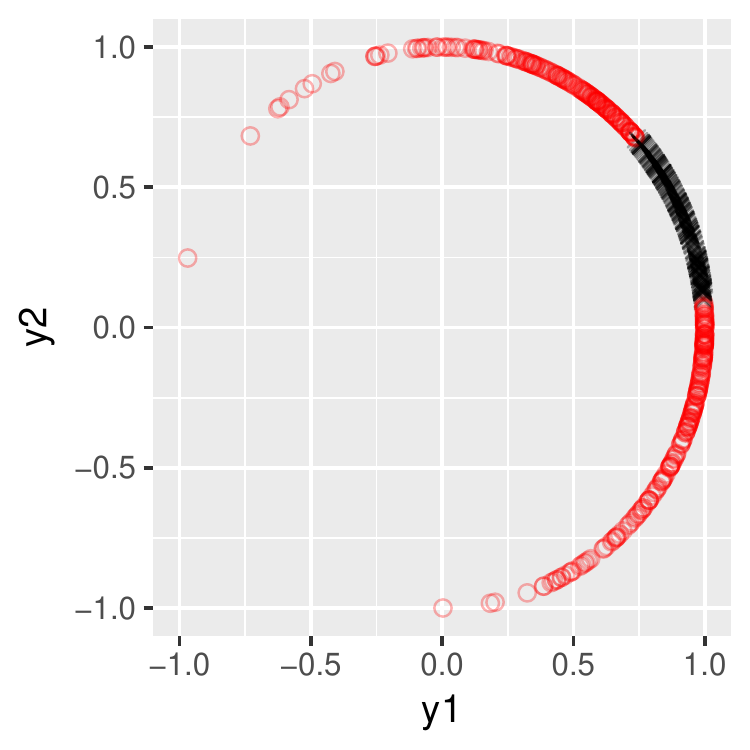}
}
\hfill
\subfloat[Point clustering estimates from Bayesian Distance Clustering.]{}{
      \includegraphics[width=1.5in, height=1.5in]{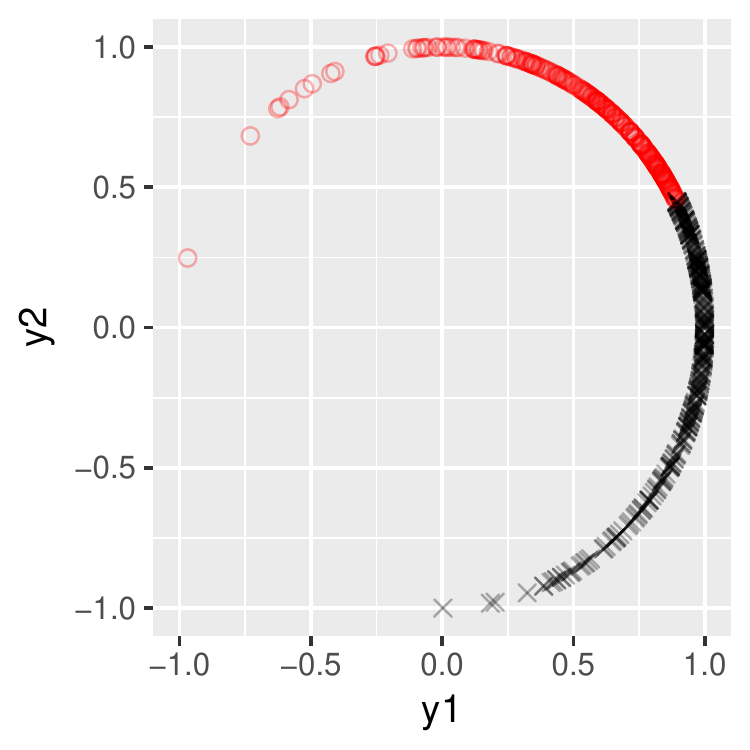}
}
\end{center}
  \caption{Clustering data from two-component mixture of von-Mises Fisher with $\boldsymbol \mu_1=(1,0)$ and $\boldsymbol \mu_2=(1/\sqrt{2},1/\sqrt{2})$. Bayesian distance clustering accurately estimates cluster labels (panel c), while mixture of Gaussians results in labels (panel b) very different from the truth.
  \label{plt:vmf_simulations}}
\end{figure}

\begin{table}
\caption{Accuracy of clustering spherical data. Adjusted Rand index (ARI) is computed for the point estimates using variation of information. The first parameter $\boldsymbol \mu_1=(1,0)$ is fixed and $\boldsymbol  \mu_2$ is chosen from $(-1,0)$, $(-\sqrt{1/5},2/\sqrt{5})$, $(1/\sqrt{2},1/\sqrt{2})$ and $(\sqrt{2/3},1/\sqrt{3})$. The average ARI and $95\%$ confidence intervals are shown.
\label{tab:sim_ari_circle}}
\centering
    \begin{tabular}{l | c c c  }
                  \hline
  $\text{Arc-length}(\boldsymbol \mu_1,\boldsymbol \mu_2)$     & Bayes Dist. Clustering  &  Mix. of Gaussians & Mix. of vMFs \\
\hline
$2$  &  1.00 (1.00, 1.00) &  1.00 (1.00, 1.00) & 1.00 (1.00, 1.00)\\
 $1.70$ &  0.65 (0.60, 0.70) &  0.60 (0.55, 0.64) & 0.65 (0.61, 0.70)\\
$0.76$   &  0.53 (0.43, 0.62) &  0.05 (0.00, 0.10) & 0.52 (0.40, 0.63)\\
$0.61$ &  0.40 (0.31, 0.45) &  0.02 (0.00, 0.05) & 0.41 (0.33, 0.45)\\
\hline
\end{tabular}
\end{table}

As shown in Table~\ref{tab:sim_ari_circle}, as the arc-length decreases, the mixture of Gaussians starts to deteriorate rapidly. This can be explained in Figure~\ref{plt:vmf_simulations}(b), where the point estimate for the mixture of Gaussians treats the heavily overlapping region as one component of small variance, and outer parts as one of larger variance. Although one could avoid this behavior by constraining Gaussian components to have the same variance, this would be sub-optimal since the variances are in fact different due to $\kappa_1\neq \kappa_2$. In contrast, Bayesian distance clustering accurately estimates clustering, as it encourages clustering data connected by small distances (Figure~\ref{plt:vmf_simulations}(c)). The result is very close to the correctly specified mixture of von Mises-Fisher distribution, as implemented in {\em Directional} package \citep{mardia2009directional}.

\section{Clustering brain regions}
We carry out a data application to segment the mouse brain according to the gene expression obtained from Allen Mouse Brain Atlas dataset \citep{lein2007genome}. Specifically, the data are {\it in situ} hybridization gene expression, represented by expression volume over spatial voxels. Each voxel is a $(200 \mu m)^3$ cube. We take the mid-coronal section of $41\times 58$ voxels. Excluding the empty ones outside the brain, they have a sample size $n=1781$. For each voxel, there are records of expression volume over 3241 different genes. To avoid the curse of dimensionality for distances, we extract the first $p=30$ principal components and use them as the source data.

Since gene expression is closely related to the functionality
of the brain, we will use the clusters to represent the functional
partitioning, and compare them in an unsupervised manner with
known anatomical regions. The voxels belong to $12$ macroscopic
anatomical regions  (Table~\ref{tb:brain_region_names}).

\begin{table}
\caption{Names and voxel counts in $12$ macroscopic anatomical
structures in the coronal section of the mouse brain. They represent
the {\em structural} partitioning of the brain.
\label{tb:brain_region_names}}
\centering
    \begin{tabular}{l l}
    \hline
    \hline
Anatomical Structure  Name               & Voxel Count\\
  \hline
  \hline
Cortical plate                & 718\\
Striatum                      & 332\\
Thalamus                      & 295\\
Midbrain                      & 229\\
Basic cell groups and regions & 96\\
Pons                          & 56\\
Vermal regions                & 22\\
Pallidum                      & 14\\
Cortical subplate             & 6\\
Hemispheric regions           & 6\\
Cerebellum                    & 5\\
Cerebral cortex               & 2\\
  \hline
\end{tabular}
\end{table}

\FloatBarrier
\begin{table}
  \captionsetup{width=.8\linewidth}
\caption{Group indices and voxel counts in $7$ clusters found by Bayesian Distance Clustering, using the gene expression volume over the coronal section of the mouse brain. They represent the {\em functional} partitioning of the brain.
 \label{tb:brain_cluster_summary}}
\centering
    \begin{tabular}{ l l  }
    \hline
        \hline
Index & Voxel Count      \\
  \hline
      \hline
1       & 626                 \\
2       & 373                  \\
3       & 176                 \\
4       & 113          \\
5       & 79             \\
6       & 39           \\
7       & 12                  \\
  \hline
\end{tabular}
\end{table}

For clustering, we use an over-fitted mixture with $k=20$ and
small Dirichlet concentration parameter $\alpha=1/20$. As shown
by \cite{rousseau2011asymptotic}, asymptotically, small $\alpha<1$
leads to automatic emptying of small clusters; we observe such
behavior here in this large sample. In the Markov chain, most
iterations have $7$ major clusters. Table~\ref{tb:brain_cluster_summary}
lists the voxel counts at $\hat {\boldsymbol c}_{(n)}$.

\begin{figure}[H]
\captionsetup[subfloat]{width=2.2in}
\begin{center}
\hspace{-1.1cm}
\subfloat[Anatomical structure labels.]{}{
      \includegraphics[width=2in, height=1.9in]{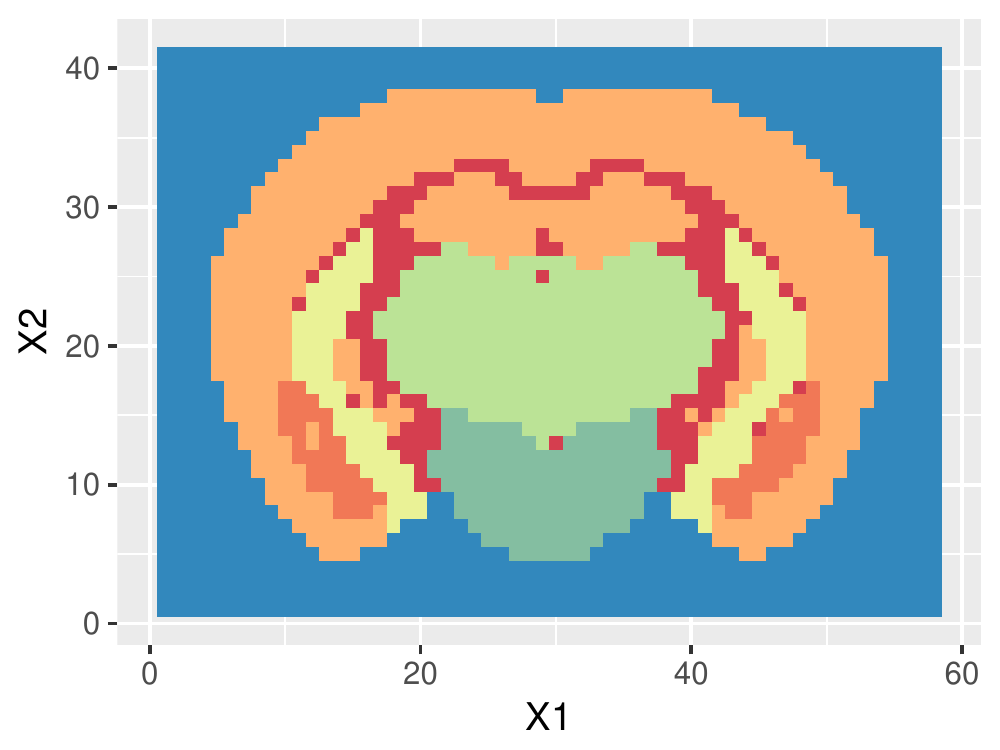}
}
\hspace{1.05cm}
\subfloat[Point estimate from Gaussian mixture model.]{}{
      \includegraphics[width=2in, height=1.9in]{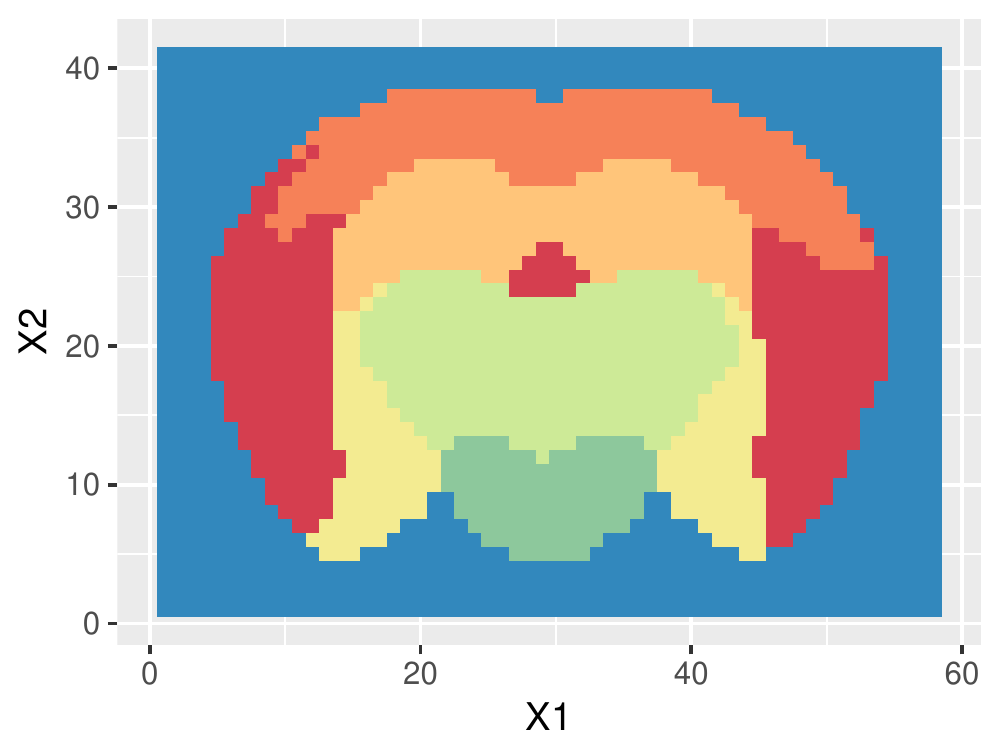}
}
\\
\subfloat[Point estimate from Bayesian Distance Clustering.]{}{
      \includegraphics[width=2in, height=1.9in]{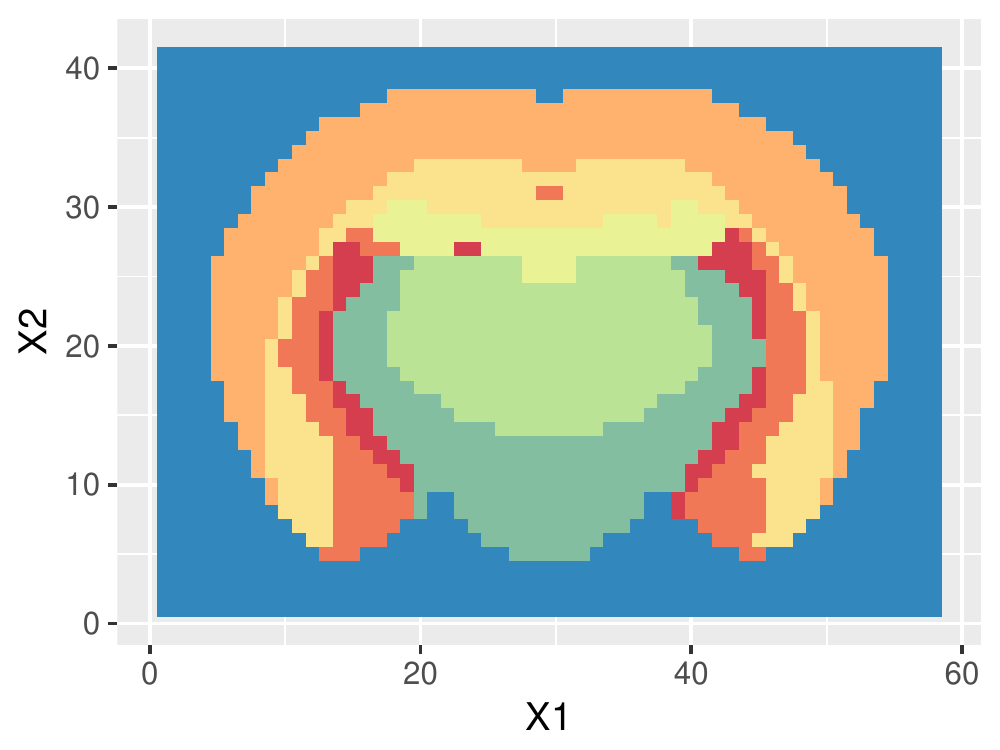}
}
\hspace{1.05cm}
\subfloat[Uncertainty based on Bayesian Distance Clustering: $\text{pr}(c_i\neq \hat c_i)$]{}{
      \includegraphics[width=2.5in, height=1.9in]{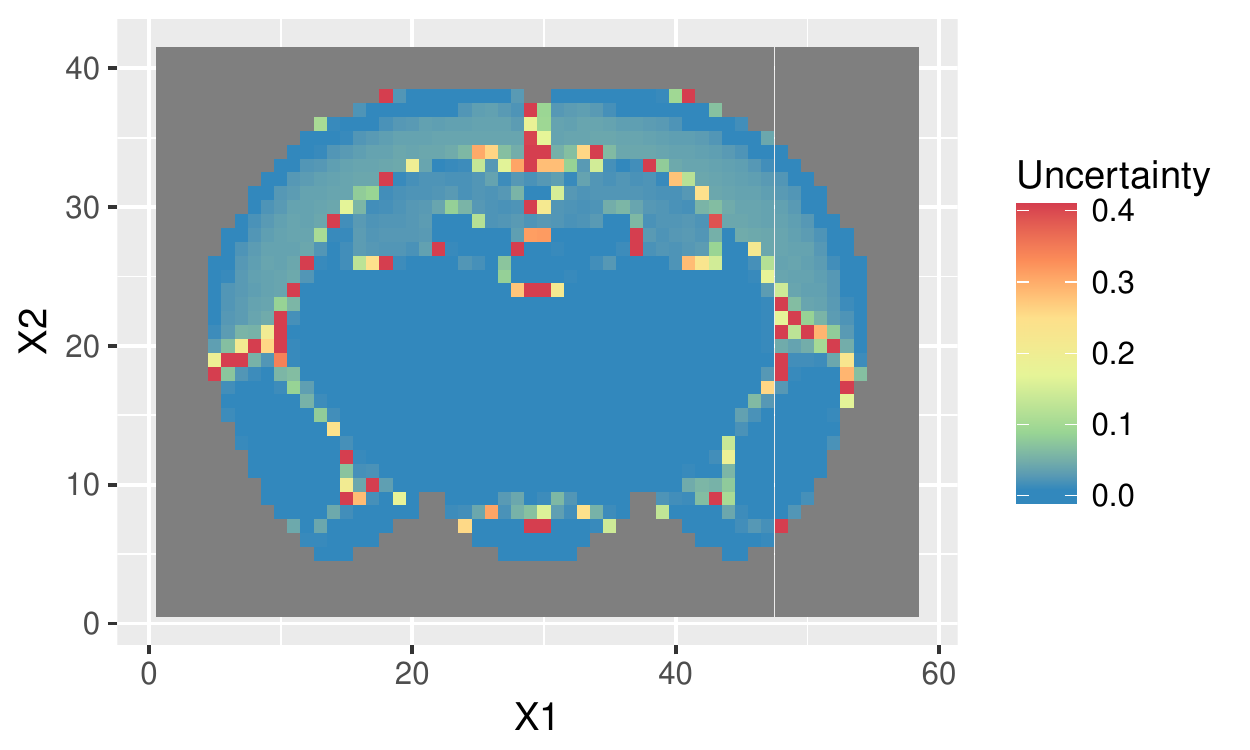}
}
\end{center}
  \caption{Clustering mouse brain using gene expression: visualizing the clustering result on the spatial grid of brain voxels. Comparing with the anatomical structure (panel a),  Bayesian Distance Clustering (panel c) has higher similarity than the  Gaussian mixture model (panel b). Most of the uncertainty (panel d) resides  in the inner layers of the cortical plate (upper parts of the brain). \label{plt:mouse_brain}}
\end{figure}

Comparing the two tables, although we do not expect a perfect
match between the structural and functional partitions, we do
see a correlation in group sizes based on the top few groups.
Indeed, visualized on the spatial grid (Figure~\ref{plt:mouse_brain}),
the point estimates from Bayesian distance clustering have very
high resemblance to the anatomical structure. Comparatively,
the clustering result from Gaussian mixture model is completely
different.

\FloatBarrier
\begin{figure}[H]
\captionsetup[subfloat]{width=1.7in}
\begin{center}
\subfloat[Anatomical structure labels.]{}{
      \includegraphics[width=1.5in, height=1.5in]{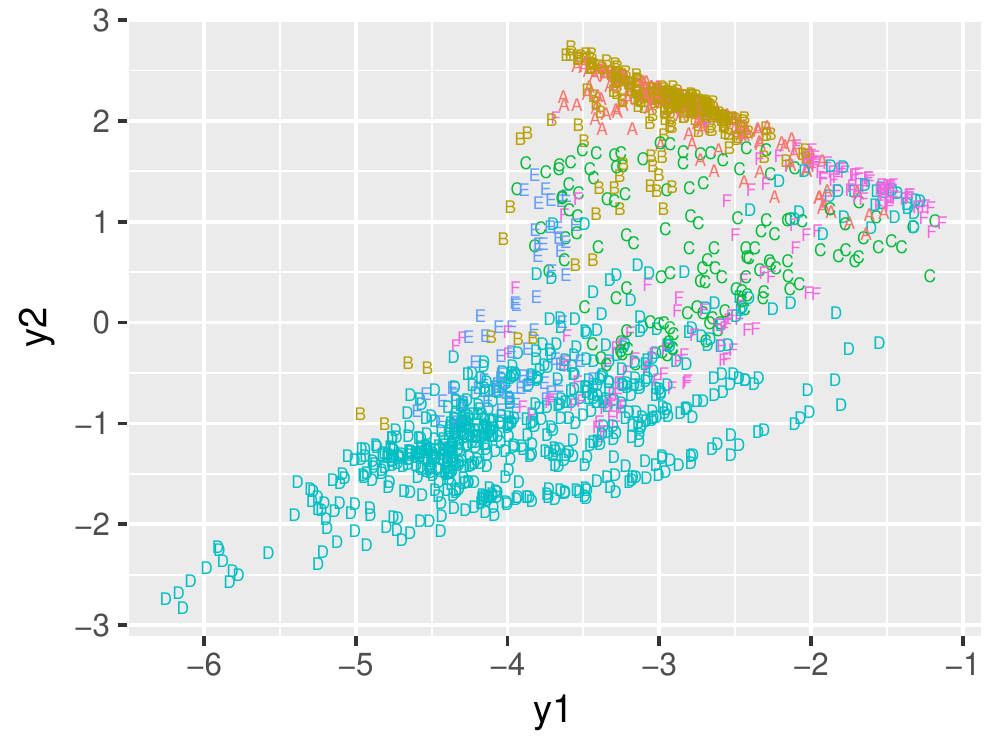}
}
 \hspace{.5cm}
\subfloat[Point estimate from Gaussian mixture model.]{}{
      \includegraphics[width=1.5in, height=1.5in]{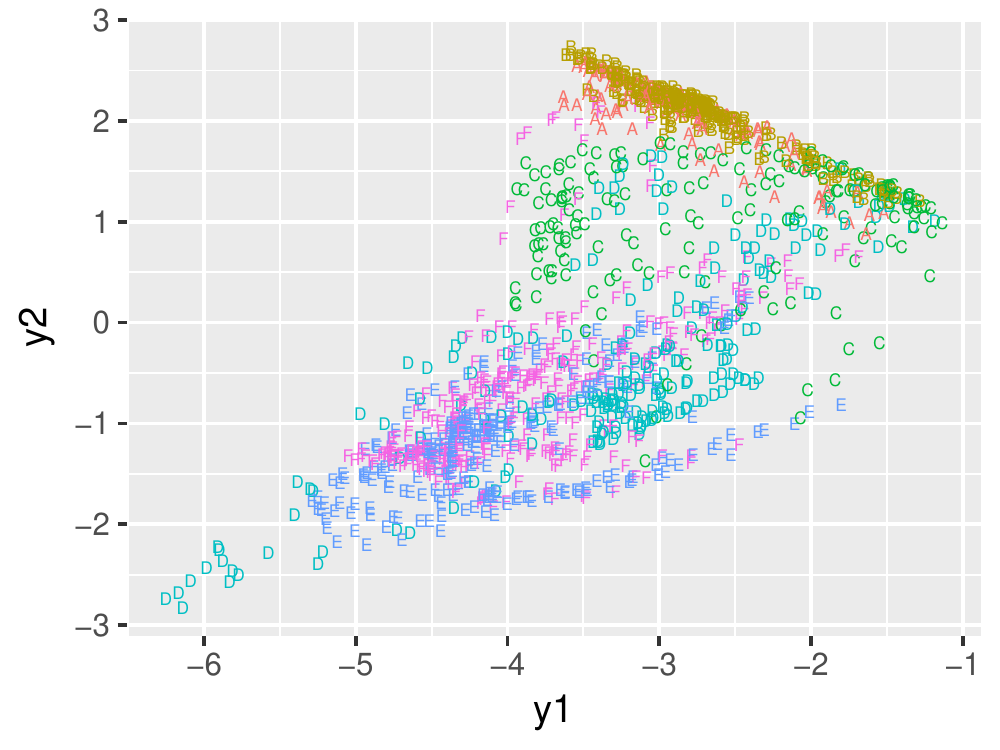}
}
 \hspace{.5cm}
\subfloat[Point estimate from Bayesian Distance Clustering.]{}{
      \includegraphics[width=1.5in, height=1.5in]{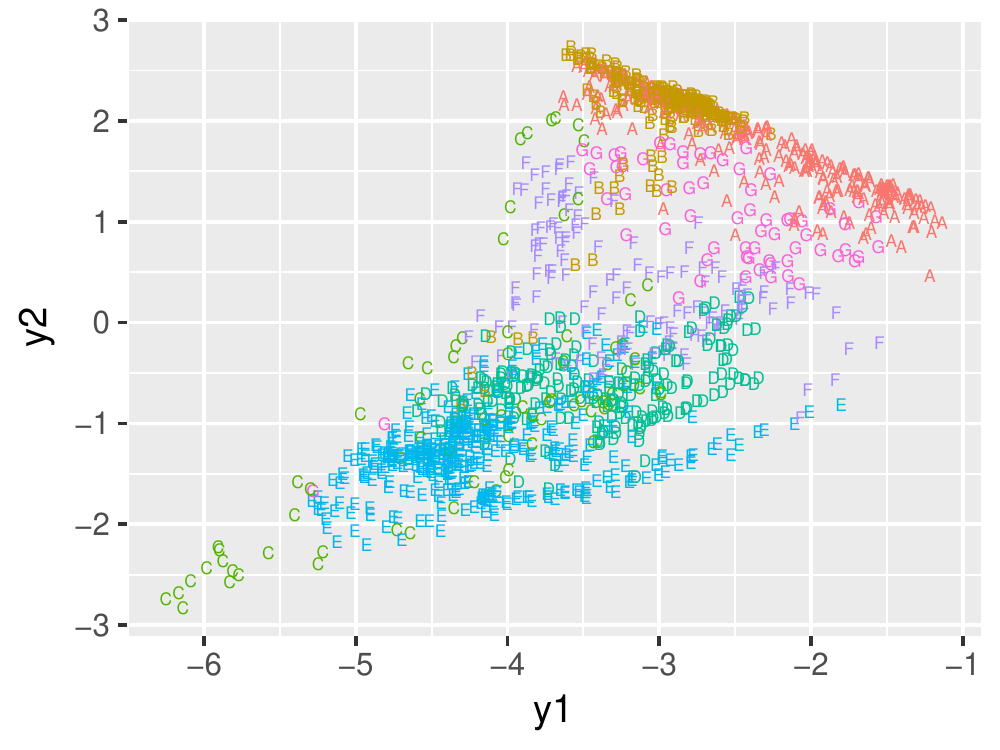}
}
\end{center}
  \caption{Clustering mouse brain using gene expression: visualizing the clustering result on the first two principal components. \label{plt:mouse_brain_pca}}
\end{figure}

\FloatBarrier
\begin{table}
\caption{Comparison of label point estimates using Bayesian distance clustering (BDC), Gaussian mixture model (GMM), spectral clustering, DBSCAN and HDClassif. The similarity measure is computed with respect to the anatomical structure labels. \label{tb:comp_point_est_brain_clust}}
\centering
\small
    \begin{tabular}{l*{8}{c}r}
    \hline
              & BDC &  GMM & Spectral Clustering & DBSCAN & HDClassif\\
\hline
Adjusted Rand Index& 0.49 & 0.31 & 0.45 & 0.43    & 0.43\\
Normalized Mutual Information & 0.51 & 0.42 & 0.46 & 0.44 & 0.47 \\
Adjusted Mutual Information & 0.51 & 0.42 & 0.47 & 0.45 &  0.47\\
    \hline
\end{tabular}
\end{table}

To benchmark against other distance clustering approaches, we compute various similarity scores and list the results in Table~\ref{tb:comp_point_est_brain_clust}.  Competing methods include spectral clustering  \citep{ng2002spectral}, DBSCAN \citep{ester1996density} and  HDClassif \citep{berge2012hdclassif}; the first two are applied on the same dimension-reduced data as used by Bayesian distance clustering, while the last one is applied directly on the high dimensional data. Among all the methods, the point estimates of Bayesian Distance Clustering have the highest similarity to the anatomical structure.

 Figure~\ref{plt:mouse_brain}(d) shows the uncertainty about the point clustering estimates, in terms of the probability $\text{pr}(c_i\neq \hat c_i)$. Besides the area connecting neighboring regions, most of the uncertainty resides in the inner layers of the cortical plate (upper parts of the brain); this is due to about $30\%$ of genes having expression concentrated only on the outer layer, leaving this part with no signals. As a result, the inner cortical plate can be either clustered with the outer layer or with the inner striatum region.

\section{Discussion}
{
 The use of a distance likelihood reduces the sensitivity to the choice of a mixture kernel, giving the ability to exploit distances for characterizing complex and structured data. While we avoid specifying the kernel, one potential weaknesses is that there can be sensitivity to the choice of the distance metrics. However, our analyses suggest that this sensitivity is often less than that to the assumed kernel. In many settings, there is a rich literature considering how to carefully choose the distance metric to reflect structure in the data \citep{pandit2011comparative}. In such cases, sensitivity of clustering results to the distance can be viewed as a positive. Clustering method necessarily rely on some notion of distances between data points.

Another issue is that we give up the ability to characterize the distribution of the original data. An interesting solution is to consider a modular modeling strategy that connects the distance clustering to a post-clustering inference model, while restricting the propagation of cluster information in one direction only. Related modular approaches have been shown to be much more robust than a single overarching full model \citep{jacob2017better}.
}

\section*{Acknowledgement}
This work was partially supported by grants R01-ES027498 and R01-MH118927 of the United States National Institutes of Health.  The authors thank Amy Herring for helpful comments.

\bibliography{reference}
\bibliographystyle{chicago}

\newpage

\appendix
\section*{Appendix}

\subsection*{Proof of Lemma 1} 
\begin{proof}
We first focus on $x\sim \text{Gamma}(\alpha,1)$,

%
By Markov's inequality
\be
\text{pr}(x\ge t) \le \frac{\mathbb{E} \exp(s X)}{\exp(st) }
=(1-s)^{-\alpha} e^{-t s},
\ee
where $s<1$. Minimizing the right hand side over $s$ yields 
 $s^*=1-\alpha/t$, and
\be
\text{pr}(x\ge t) \le (\frac{t}{\alpha})^{\alpha} e^{-t +\alpha}
= \alpha^{-\alpha} e^\alpha t^\alpha e^{-t}.
\ee
Scaling $x$ by $\sigma_h$ and adjusting the constant yield the
results.

\end{proof}

\subsection*{Proof of Theorem 1}

\begin{proof}
Equivalently, sub-exponential tail can be characterized by the
bound on its moment generating function
\begin{equation*}
        \begin{aligned}
\mathbb{E} \exp\{ t( y^{[h]}_{i,j}-\mu^{[h]}_j)\}\le \exp({\nu_h^2
t^2/2}) \quad \forall  |t|\le 1/b_h,
        \end{aligned}
\end{equation*}
for $j=1,\ldots,p$.
It immediately follows that the pairwise difference $\tilde{\boldsymbol d}^{[h]}_{i,i'}=\yh{i}-\yh{i'}$
between two iid random variables must be sub-exponential as well,
with
\begin{equation*}
        \begin{aligned}
\mathbb{E} \exp( t \tilde d^{[h]}_{i,i',j})\le \exp({\nu_h^2
t^2}) \quad \forall  |t|\le 1/b_h.
        \end{aligned}
\end{equation*}
Then the vector norm
\begin{equation*}
        \begin{aligned}
                \text{pr}( \dh{ii'} > p^{\eta} t) &= \text{pr}
( \sum_{j=1}^p |\tilde d^{[h]}_{ii',j}|^q>p^{\eta q}t^q) \\
                                        & \le p \; \text{pr}
( |\tilde d^{[h]}_{ii',j}|^q> p^{\eta q -1}t^q) \\
                                        & =  p \; \text{pr} (
|\tilde d^{[h]}_{ii',j}|> p^{\eta  -1/q}t)\\ 
                                        & \le 2 p \exp\{-t p^{(\eta-1/q)}/(2b_h
)\}  \quad \text{  for  } t >  p^{1/q-\eta } 2\nu_h^2 /b_h.
        \end{aligned}
\end{equation*}
where the first inequality is due to $\text{pr}(\sum_{i=1}^{p}
a_i> b)\le p \sum_{i=1}^p\text{pr}( a_i> b/p) $ and second inequality
uses the property of sub-exponential tail \citep{wainwright2019high}.
\end{proof}

%
\subsection*{Proof of Theorem 3}
\begin{proof}
For a clear exposition, we omit the sub/super-script $h$ for now and use $\boldsymbol x_i =T(\boldsymbol y_i)$
    \begin{equation*}
        \begin{aligned}
\mathbb{E}_{\boldsymbol y_i} \mathbb{E}_{\boldsymbol y_j}     \sum_{i=1}^{n}\sum_{j=1}^{n} B_\phi(\boldsymbol x_i,\boldsymbol x_j) =  & \mathbb{E}_{\boldsymbol y_i} \mathbb{E}_{\boldsymbol y_{j}}\sum_{i=1}^{n}\sum_{j=1}^{n} \{\phi(\boldsymbol x_i) - \phi(\boldsymbol x_j) - \langle\,\boldsymbol x_i-\boldsymbol x_j, \triangledown \phi(\boldsymbol x_j)\rangle \}   \\
      =  & \mathbb{E}_{\boldsymbol y_j}\sum_{j=1}^{n}  \sum_{i=1}^{n}\{ \mathbb{E}_{\boldsymbol y_i}\phi(\boldsymbol x_i) -  \phi(\boldsymbol\mu)  -  \langle\, \mathbb{E}_{\boldsymbol y_i} \boldsymbol x_i-\boldsymbol \mu, \triangledown \phi(\boldsymbol\mu)\rangle \\
    &  + \phi(\boldsymbol\mu) -  \phi(\boldsymbol x_j) - \langle\,\mathbb{E}_{\boldsymbol y_i}\boldsymbol x_i-\boldsymbol x_j, \triangledown \phi(\boldsymbol x_j)\rangle   \\
        =  & n \sum_{i=1}^{n}\mathbb{E}_{\boldsymbol y_i}\{\phi(\boldsymbol x_i) -  \phi(\boldsymbol\mu)  -  \langle\,\boldsymbol x_i- \boldsymbol\mu, \triangledown \phi(\boldsymbol\mu)\rangle \}\\
    &  + n \sum_{j=1}^{n}  \mathbb{E}_{\boldsymbol y_j}\{  \phi(\boldsymbol\mu) - \phi(\boldsymbol x_j) - \langle\, \boldsymbol\mu- \boldsymbol x_j, \triangledown \phi(\boldsymbol x_j)\rangle \}   \\
=&n \sum_{i =1}^{n} \mathbb{E}_{\boldsymbol y}\{B_\phi(\boldsymbol x_i, \boldsymbol\mu) + B_\phi(\boldsymbol\mu,\boldsymbol x_i) \},  
        \end{aligned}
\end{equation*}
where $\langle .,. \rangle $ denotes dot product, the second equality is due to  Fubini theorem and $ \mathbb{E}_{\boldsymbol y_i}\boldsymbol x_i -\boldsymbol  \mu= 0$.
\end{proof}

\subsection*{Proof of Theorem 4}

\begin{proof}
Using ${\bf 1}_{n,m}$ $n\times m$ matrix with all elements equal
$1$.
        Since $C^{\rm T} C = \text{diag}(n_1,\ldots,n_k)$, the
2 times of normalized graph cut loss can be written as
\begin{equation*}
\begin{aligned}
& \text{tr} \big[ A ({\bf 1}_{n,k}- C)   \big (C^{\rm T} C  \big)^{-1}
C^{\rm T} \big] \\
  & =
 - \text{tr} \big\{   A C   \big (C^{\rm T} C  \big)^{-1} C^{\rm
T}\big\}  +  \text{tr} \big[A  {\bf 1}_{n,k}
  \big (C^{\rm T} C  \big)^{-1} C^{\rm T} \big].
  \end{aligned}
\end{equation*}
For the second term
\be
& \text{tr} \big[A  {\bf 1}_{n,k}
  \big (C^{\rm T} C  \big)^{-1} C^{\rm T} \big] \\
=  & \text{tr} \big[ C^{\rm T} A {\bf 1}_{n,k}
  \big (C^{\rm T} C  \big)^{-1}\big] \\
=&      
 \sum_{i=1}^n \frac{ \sum_{j=1}^nA_{i,j}}{n_{c_i}}.
 \ee
\end{proof}

\end{document}